%% file: main.tex
\documentclass{article}
\usepackage{spconf,amsmath,graphicx,hyperref}
\usepackage{amsfonts,xcolor,comment,pifont,tipa,booktabs}
\usepackage[whole]{bxcjkjatype}

\definecolor{lightgrey}{rgb}{0.925, 0.925, 0.925}

\newcommand{\token}[1]{\texttt{#1}}
\newcommand{\Fzero}{ $f_{o}$ }
\newcommand{\boldFzero}{ $\boldsymbol{f_{o}}$ }
\newcommand{\fst}[2]{\mathcal{#1}^{\mathrm{#2}}}

\title{BUILDING TAILORED SPEECH RECOGNIZERS FOR JAPANESE SPEAKING ASSESSMENT}

\name{Yotaro Kubo, Richard Sproat, Chihiro Taguchi, Llion Jones}
\address{Sakana AI, Tokyo, Japan.\\
\texttt{\{yotarokubo,rws,chihirotaguchi,llion\}@sakana.ai}
}

\begin{document}
\ninept
\maketitle

\begin{abstract}
This paper presents methods for building speech recognizers tailored for Japanese speaking assessment tasks.
Specifically, we build a speech recognizer that outputs phonemic labels with accent markers.
Although Japanese is resource-rich, there is only a small amount of data for training models to produce accurate phonemic transcriptions that include accent marks.
We propose two methods to mitigate data sparsity. First, a multitask training scheme introduces auxiliary loss functions to estimate orthographic text labels and pitch patterns of the input signal, so that utterances with only orthographic annotations can be leveraged in training. The second fuses two estimators, one over phonetic alphabet strings, and the other over text token sequences. To combine these estimates we develop an algorithm based on the finite-state transducer framework.
Our results indicate that the use of multitask learning and fusion is effective for building an accurate phonemic recognizer.
We show that this approach is advantageous compared to the use of generic multilingual recognizers.
The relative advantages of the proposed methods were also compared.
Our proposed methods reduced the average of mora-label error rates from 12.3\% to 7.1\% over the CSJ core evaluation sets.
\end{abstract}

\input{introduction}

\input{system_design}

\input{multi_task_learning}

\input{lattice_fusion}

\input{experiments}

\input{conclusions}

\vfill\pagebreak

\bibliographystyle{IEEEbib}
\bibliography{strings,refs}

\end{document}

%% file: introduction.tex
\section{Introduction}

Automatic speech recognition (ASR) technology has been evolving to to enable the mapping of speech signals to canonical textual representations under a variety of conditions.
In typical ASR systems, speaker errors, such as incorrect accent positions, mispronunciations, fillers, and hesitations, are intentionally discarded.
Although many applications prefer such canonical output rather than a phonetically accurate transcription, in certain cases such as language education, this normalization effect can be viewed as a limitation because it hinders the accurate evaluation of users' speaking proficiency.

There have been several attempts to build universal phonetic transcribers, with a particular focus on extending ASR to low-resource languages \cite{li2020universal,xu2022simple,taguchi2023universal}.
While multilingual models are potentially applicable to speaking skill assessment, assessing language-specific phenomena, such as pitch accent, is also essential for educational purposes.
In this paper, we focus on phonemic transcription of Japanese along with pitch accent markers, for the purposes of building a speech recognizer tailored for Japanese speaking assessment.
Considering the complexity and irregularity of Japanese pitch accent rules \cite{sagisaka1984prosodic,kubozono1987}, this presents a unique challenge that cannot be solved only by multilingual phonetic recognizers. %

The training of phonetically accurate recognizers often suffers from the limited availability of accurate phonetic or phonemic transcriptions.
Although Japanese is a relatively resource-rich language, the largest multi-speaker dataset with hand-annotated pitch accent is, to the best of our knowledge, the ``core'' subset of the Corpus of Spontaneous Japanese (CSJ) \cite{maekawa2003corpus}, containing only 45 hours of speech.

Prior work by \cite{ohnaka2025grapheme} developed a phonemic recognizer conditioned on the outputs of external speech recognizers.
Although their method did not include a built-in ASR component, there are several similarities to our proposal.
Their system performs both implicit conditioning, employing text tokens as an auxiliary input, and explicit conditioning using text information as constraints on decoding.
Unlike their implicit conditioning method, our approach does not require reliable text labels because it solves text and phonetic alphabet (PA) recognition simultaneously in a multitask learning framework.
Further, as discussed in the following section, one can view our decoding method as an extension of their explicit conditioning method that can robustly handle weak text constraints.
For a conventional ASR task, the iterative refinement method is proposed in \cite{kubo2020joint} in order to estimate PA and text token sequences jointly.
However, this method does not estimate accent locations, and does not utilize a dictionary to enhance accuracy in small data scenarios.

The technical contributions in this paper are twofold: First, a multitask training method that is effective for pitch-accent recognition; second, a decoding method that decodes an optimal PA sequence robustly by considering both text tokens and PA estimation results.
These two novel methods are implemented in a streamable ASR model, and comparative experiments were conducted to confirm the effectiveness of the proposed methods.
Fig.~\ref{fig:diagram} illustrates our proposed method. The next three sections describe the components in the figure.

%% file: system_design.tex
\section{System Design}

\begin{figure}[bt]
  \centering
\centerline{\includegraphics[width=1.0\linewidth]{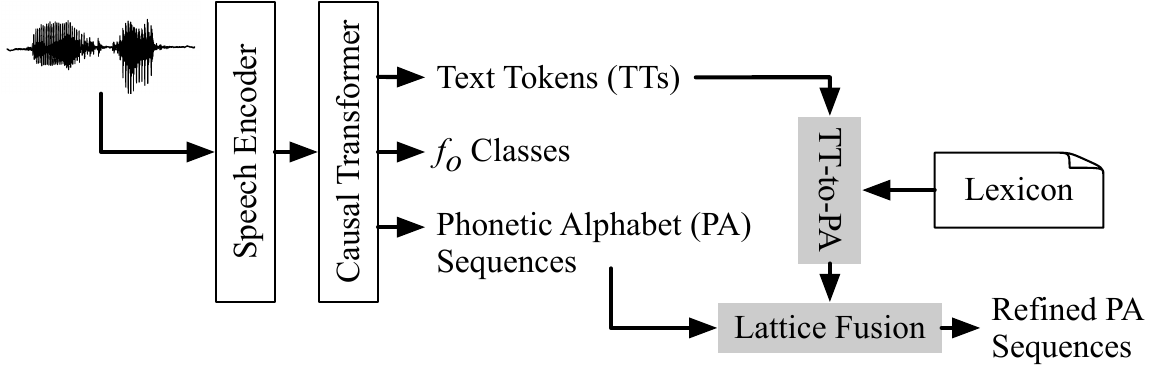}}
\caption{\small Proposed architecture of our PA recognizer. Boxes with white backgrounds are neural components, and those with  dark backgrounds are algorithmic processing.}
\label{fig:diagram}
\vspace*{-5mm}
\end{figure}

\subsection{Connectionist Temporal Classification}

Conventionally, language models are integrated into ASR systems in order to improve the output quality leveraging the frequencies of phrases and words.
However, since one of our goals is to detect errors in speech that manifest as infrequent fluctuations in token sequences, such corrections by language models may conflict with this objective.

Connectionist Temporal Classification (CTC) can be a solution to this problem.
Unlike other ASR architectures, such as RNN-T \cite{graves2012sequence} and LAS \cite{chan2016listen}, a CTC model does not include a module for capturing mutual dependency within the output sequences.
Therefore, it is expected that CTC models are incapable of correcting erroneous utterances into canonical ones.
While it has been reported that even CTC can implicitly learn an internal language model \cite{zhao2025regarding,yang2025label}, CTC is still considered the best choice if correction needs to be minimized.

\subsection{Corpus of Spontaneous Japanese (CSJ)}

For our purposes, a training set annotated with PAs and accent markers is required.
The transcription in the dataset must be faithful to the original speech, including errors.
CSJ includes manual annotations for such speech variation. By training on this dataset, one can obtain a recognizer that can detect such errors represented in PAs.
A subset of CSJ (called ``core'') provides mora-wise phonemic annotations in katakana and the perceived location of accent;\footnote{Japanese is a \emph{mora-timed} language, a mora being a prosodic unit corresponding to a short vowel (V) or a single post-vocalic consonant (C) (Syllable-initial consonants do not count for moraic purposes).
Syllables of the form (C)V are monomoraic, and those
of the form (C)VV or (C)VC are bimoraic.
Each mora ideally occupies about the same amount of time, though see \cite{warner2001} for a detailed review of the situation for Japanese.}
however, this subset is only a small fraction of the entire dataset,
consisting of only 23,683 training utterances, or less than 6\% of the entire CSJ training dataset.
Although Japanese is considered to be a resource-rich language, given the sparsity of accent-annotated datasets, it is still not straightforward to build an accurate recognizer out of the available data.
The next sections describe practical remedies for this limitation.

%% file: multi_task_learning.tex
\section{Multitask learning}

The sparsity of phonetic labels with accent information is the central challenge in developing speech recognizers tailored for Japanese speaking assessment.
The first remedy we employ is multitask learning.
Orthographic transcription is cheaper to obtain compared to phonemic annotation, especially if the phonemic annotation also requires accent annotation.
Since text transcriptions and phonemic annotations are expected to be mutually dependent, solving two tasks simultaneously can help learning a common representation of speech.

On the other hand, estimated pitch information is also cheap to obtain via a high-quality fundamental frequency (\Fzero) extraction algorithm.
Since Japanese has lexical pitch accent, pitch information is also expected to be highly correlated with the target labels.
Thus, in this paper, we train a transformer with three estimation tasks: one for PAs, one for text tokens (TTs), and one for estimated \Fzero classes.

\subsection{Encoder Description}

To maximize downstream utility, the proposed model is designed to be a streamable model.
In order to ensure streamability, we adopt a streamable speech encoder and a causal transformer.
For the speech encoder, we adopt the Mimi \cite{kyutai2024moshi} model pretrained with 7 million hours of multilingual (mostly English) datasets \cite{kyutai_mimi_pretrained}.
Since our model does not require discrete inputs, the quantization and down-sampling modules are removed from Mimi.

Our transformer model is based on Llama-2 \cite{touvron2023llama}.
Although Llama-2 was originally proposed for auto-regressive decoder-only modeling, our method adopts this architecture for making an encoder-only CTC model.
Model parameters are randomly initialized.
The embedding dimensionality is set to 512, the number of layers to 24,
and the number of attention heads to 8.
To overcome overfitting, dropout is introduced in several layers.
A dropout with probability of 0.2 is applied to the input embeddings from the Mimi encoder and the attention probabilities in each self-attention layer.

\subsection{Phonetic Alphabet and Text Token Tasks}

Our PA tokenizer assumes a katakana transcription of the utterance.
If an accent is placed on a mora, this is denoted by appending an apostrophe to the mora label (e.g. \token{キュ} /\textipa{k\super jW}/ with accent is denoted as \token{キュ'}).
The long-vowel marker (\token{ー}) is replaced by a copy of the preceding vowel. 
For example, the \token{キュ}\token{ー} /\textipa{k\super jW\textlengthmark}/ is replaced by \token{キュ}\token{ウ} /\textipa{k\super jW W}/.
Similarly, \token{キュ'}\token{ー} is replaced by \token{キュ'}\token{ウ}.
We collect all tokens from the CSJ core training set, resulting in 243 unique output tokens.

Text tokens (TTs) are character-based. We collected all Unicode points from the CSJ training dataset (core and non-core) after NFKC normalization, resulting in 2,309 distinct tokens.

We use two CTC estimators for PAs and TTs, each containing a fully-connected layer for computing logits of the CTC label distribution---i.e. the TT or PA token plus a ``blank'' token indicating absence of the corresponding output label.

\subsection{\texorpdfstring{\Fzero}{F0} Classifier Task}

The objective of the \Fzero classifier task is to ensure that pitch information is propagated through the transformer, allowing the model to estimate \Fzero contours.
For each utterance, the Harvest algorithm\cite{morise2017harvest} is applied to estimate \Fzero each 10 ms.
Since pitch accent is based not on the pitch itself, but rather its trajectory, the task is designed to estimate predefined classes that indicate if \Fzero is going up or down.

For defining classes over \Fzero patterns, three analysis windows are introduced for each input frame.
Let $t_n$ be the timestamp of the $n$-th input frame for the transformer.
The time segments for left, central, and right analysis windows are defined as $L_n = [t_n - 1.5 w, t_n - 0.5w)$, $C_n = [t_n - 0.5w, t_n + 0.5w)$, and $R_n = [t_n + 0.5w, t_n + 1.5w)$, respectively, where $w$ is the window length set to 40 ms.
For each analysis window, we compute voicedness $V(S) \in \{ \mathtt{voiced}, \mathtt{unvoiced} \} $, where $S$ is either $L_n$, $C_n$ or $R_n$. If there is a valid \Fzero estimation in the time segment $S$, $V(S) = \mathtt{voiced}$, and $V(S) = \mathtt{unvoiced}$ otherwise. The average of log-\Fzero $P(S)$ is computed for the voiced time segment $S$, i.e. $V(S) = \mathtt{voiced}$.

The number of possible combinations of $V(L_n)$ and $V(R_n)$ is 4.
For the case where $V(L_n) = V(R_n) = \mathtt{voiced}$, we can further split the class depending on whether 
${\rm log}$-$f_{o}$
is going up or not, i.e. $P(L_n) < P(R_n)$.
Therefore, we can classify the left $L_n$ and right $R_n$ windows into 5 classes, which can be further split by the voicedness of the central window $V(C_n)$.
In the end, we classify a frame into 10 classes based on the estimated \Fzero trajectory.

A fully connected layer is introduced to estimate the logits over the \Fzero classes.
The training loss is enhanced by adding a term representing the cross-entropy loss of this class estimator.

%% file: lattice_fusion.tex
\section{Lattice Fusion}

\begin{figure}[bt]
  \centering
  \centerline{\includegraphics[width=1.0\columnwidth]{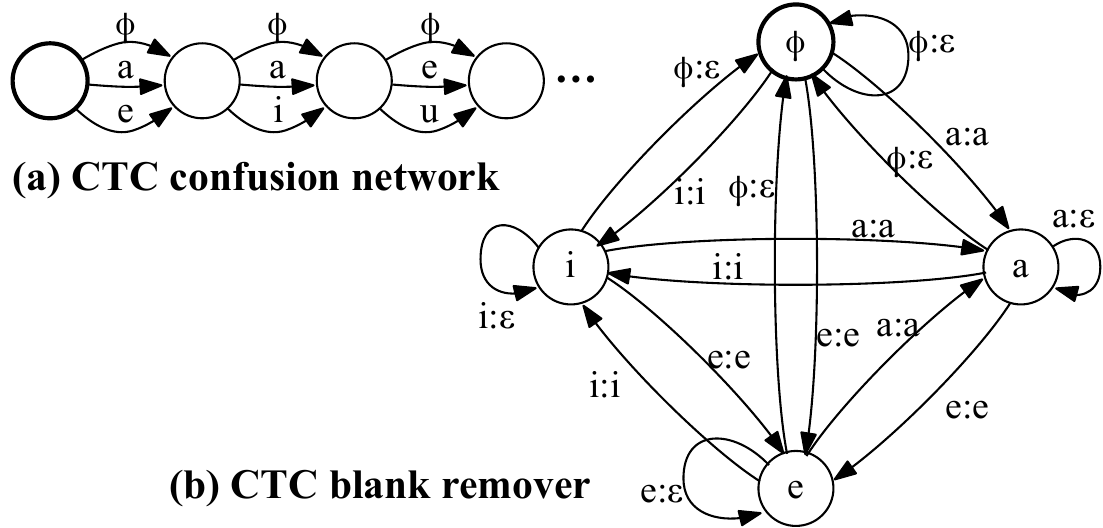}}
\caption{\small Finite-state representation of (a) CTC outputs and (b) blank removal. For brevity, only phonemes /a, i, e/ and blank $\phi$ are shown.}
\label{fig:fsts}
\vspace*{-5mm}
\end{figure}

To increase the accuracy of the PA recognizer in the face of data sparsity, our method consolidates estimation results of TT and PA recognition using finite-state transducers (FST).
Whereas finite-state acceptors (FSA) represent probabilistic distributions over sequences, FSTs represent probabilistic mappings between sequences.
We use FSAs---conventionally called ``lattices''---to represent distributions over PA and TT sequences.
The TT lattice is converted to a second PA lattice using an FST constructed from a pronunciation dictionary.
Finally, the most plausible PA sequence from the union of the two PA lattices is extracted. We term this procedure  ``lattice fusion''.

\subsection{Lattice Construction from CTC outputs}

Let $Y \in \mathbb{R}^{T \times K}$ be an array of log probabilities of each class in CTC output layer, including the special blank symbol $\phi$, where $y_{t, k}$ denotes the log-probability of label $k$ emitted at time $t$.
For each utterance, a confusion network over the CTC labels can be constructed as in Fig.~\ref{fig:fsts}~(a).
In confusion networks, each non-final state represents the timestamp $t$ and outgoing arcs from state $t$ are defined for each possible $k$. The arc labeled as $k$ from state $t$ is designed to have a weight representing the negative log probability as $w = - y_{t, k}$.
Let $\fst{S}{P}$ be the confusion network constructed by this procedure.

The CTC output labels must be post-processed by removing the same-label repetition and the blank $\phi$ symbols.
This post-processing can be represented as a transducer as in Fig.~\ref{fig:fsts}~(b). By applying the composition operator to the confusion network $\fst{S}{P}$ and the CTC postprocessing FST $\fst{B}{P}$, and projecting the FST to the ouput labels, the lattice $\fst{L}{P}$ that represents a probability distribution over all possible PA sequences can be computed.
For computational efficiency, we further applied a lattice pruning method and further optimization for maximizing the efficiency of the subsequent procedures, as follows:
 $\fst{L}{P} = \mathrm{Opt}( \mathrm{\pi_O}( \fst{S}{P} \circ \fst{B}{P} ))$.
 Here, $\circ$ denotes the composition operator on FSTs, $\mathrm{\pi_O}$ is a projection function that obtains an FSA from the FST by removing its input labels, and $\mathrm{Opt}$ is an optimization procedure that prunes, removes  $\epsilon$, determinizes, and minimizes \cite{mohri1997finite} the input, in this order, over the log semiring.
 Following the procedure, a compact representation of PA distribution can be obtained from the output log-probabilities of the PA-CTC task.

We can similarly obtain TT sequence distributions from log-probabilities in the TT-CTC task.
Let $\fst{S}{T}$ be the confusion network constructed in the same way from the TT log-probabilities.
Following the above procedure, a lattice for TTs $\fst{L}{T}$ can be obtained.

\subsection{Conversion from TT lattice to PA lattice}

Our TT-to-PA FST is based on UniDic \cite{den-etal-2008-proper}.
Of 876,803 UniDic entries, 873,647 were selected by discarding entries with null pronunciations or unknown PAs.
Lexicon FST construction followed \cite{mohri1997finite}.
Using the TT-to-PA converter represented as an FST, a lattice representing distribution over PA sequences can also be obtained by converting the TT lattice to a PA lattice.
This conversion can be done via FST composition.
Let $\fst{D}{}$ be an TT-to-PA FST, a PA lattice induced by the TT lattice can be computed as: $\fst{L'}{T2P} = \mathrm{Opt}(\mathrm{\pi_O}(\fst{L}{T} \circ \fst{D}{}))$.
Furthermore, since the dictionary $\fst{D}{}$ does not provide relative probabilities when a word has multiple pronunciation, it is necessary to accumulate PA weights to $\fst{L'}{T2P}$.
The PA probability can be taken from $\fst{L}{P}$.
The pronunciation-weighted lattice obtained in this way can be expressed as:
$\fst{L}{T2P} = \mathrm{Norm}(\fst{L}{P} \circ \fst{L'}{T2P})$.
Here, $\mathrm{Norm}$ is the normalization operator that normalizes the weights of outgoing arcs for each state to satisfy the sum-to-one constraint.

The explicit conditioning method proposed in \cite{ohnaka2025grapheme} is equivalent to selecting the most plausible path from $\fst{L}{T2P}$ obtained as above.
Unlike the explicit conditioning method, our method aims at robust decoding by fusing the estimation results from $\fst{L}{T2P}$ and $\fst{L}{P}$.
Fusing the two estimators can simply be performed by the union operator followed by optimization.
The union lattice $\hat{\fst{L}{}}$ can be computed as:
$\hat{\fst{L}{}} = \mathrm{Opt}(\fst{L}{P} \uplus \fst{L}{T2P})$,
where $\uplus$ denotes the FST union operator.
$\hat{\fst{L}{}}$ corresponds to the average of the two probabilistic distributions represented by $\fst{L}{P}$ and $\fst{L}{T2P}$.
The final recognition results can be obtained by computing the shortest path over the union lattice.

%% file: experiments.tex
\section{Experiments}

\subsection{Experimental Setup}

For analyzing the performances of the proposed systems, we compared three variants of the method: the full proposed method (MT+LF), a variant without lattice fusion (MT), and a variant with neither lattice fusion nor multitask learning (PA-only).
Further, instead of lattice fusion, we implemented explicit conditioning \cite{ohnaka2025grapheme} using our recognizers TT and PA output results (MT+Cond.).
For LF systems, the MT+LF system computed the TT lattices using the TT outputs of the MT model.
We additionally compared LF method with other sources of TT lattices. Whisper, and our ``TT-only'' variant were used as external TT lattice sources. For LF with Whisper, the lattice was constructed to have only a single path corresponding to the recognition result from Whisper.
As side results, we also compare character error rates (CERs) of the MT model, Whisper, and our model without PA and \Fzero outputs (TT-only).

For training, the CSJ dataset is augmented with speed perturbation \cite{ko15_interspeech} followed by a time-domain variant of SpecAugment \cite{park2019specaugment}.
For speed perturbation, 20\% of samples were modified to have 90\% rate, and another 20\% of samples were modified to have 110\% rate.
The time-domain SpecAugment is developed to apply masking as in the original SpecAugment but in the raw-waveform domain.
This variant implements time-masking on the raw waveform, and the spectral masking on the (full utterance) discrete Fourier transform domain.
The hyper-parameters for SpecAugment were set as follows: The number of time-domain masks was $10$, the number of frequency-domain masks was $2$, the maximum length of time-domain masks was $0.05 T$ where $T$ is the length of the input utterance, and the maximum length of frequency domain mask was $0.3 F$, where $F$ is the bandwidth of the input signal in mel.
The task weights were preliminarily set to 0.3, 0.6, and 0.1 for the PA, TT, and \Fzero tasks respectively. We did not tune the task weights.

As baseline systems, we adapted Whisper \cite{radford2023robust} (\texttt{whisper-1} model from OpenAI) and Multipa \cite{taguchi2023universal}. Since neither system is able to output our target PA directly, we made adapters that produce multiple PA outputs corresponding to the Whisper output text, or Multipa output IPAs. The minimum error rates among all possible outputs from the adapter were used as a final metric for the systems with such adapters. For the Whisper adapter, since our adapter is based on UniDic, the adapter can also output PAs with accent markers based on the accent annotation given for each word in UniDic.

For evaluation, two datasets with accent annotations were employed.
Since our model training is dependent on CSJ, core subsets of CSJ evaluation sets were used as our primary evaluation sets (CSJ eval1/ eval2/ eval3). In addition to CSJ, ``basic5000'' from the JSUT corpora was used \cite{sonobe2017jsut}. This dataset (JSUT) contains read utterances from a single speaker.
CSJ's spontaneous and multi-speaker properties align more to our objective; however, JSUT is also important for probing the behavior of our methods in out-of-domain settings.

\subsection{Discussion}

\begin{table}[t]
\vspace{-6pt}%
\caption{\small Mora-label error rates [\%] of PA recognition. (${}^\dagger =$ ASR output has multiple possible katakana representation, and minimum error rates are shown here. See main text for details.)}
\label{table:mler}
\centering
\footnotesize
\begin{tabular}{lc@{\hskip 1\tabcolsep}cc@{\hskip 1\tabcolsep}cc@{\hskip 1\tabcolsep}cc@{\hskip 1\tabcolsep}c}
\toprule
& \multicolumn{2}{c}{eval1} & \multicolumn{2}{c}{eval2} & \multicolumn{2}{c}{eval3} & \multicolumn{2}{c}{JSUT} \\
accent err. & & \ding{51}  & & \ding{51}  & & \ding{51} & & \ding{51} \\
\midrule
\textbf{Whisper}${}^{\dagger}$ & 20.6 & 24.3 & 15.1 & 19.7 & 13.5 & 16.5 & 2.8 & 6.8 \\
\textbf{Multipa}${}^{\dagger}$ & 17.2 & -- & 19.8 & -- & 18.3 & -- & 16.2 & -- \\
\midrule
\textbf{PA-only} & 7.2 & 11.5 & 7.6 & 11.9 & 7.9 & 13.5 & 6.6 & 13.2 \\
\textbf{MT} & 4.4 & 7.3 & 4.9 & 7.8 & 5.0 & {\bf 9.1} & {\bf 5.8} & {\bf 9.9} \\
\textbf{MT+Cond.} & 4.3 & 7.3 & 5.2 & 8.2 & 5.0 & 9.3 & 6.1 & 10.3 \\
\textbf{MT+LF} & {\bf 4.1} & {\bf 7.0} & {\bf 4.8} & {\bf 7.7} & {\bf 4.9} & {\bf 9.1} & 6.1 & 10.3 \\
\midrule
\multicolumn{9}{l}{\textbf{MT+LF} with external ASR models} \\
with Whisper & 4.3 & 7.2 & 4.8 & 7.6 & 4.6 & 8.8 & {\bf 2.1} & {\bf 6.4} \\
with TT-only & {\bf 3.1} & {\bf 6.0} & {\bf 4.1} & {\bf 7.1} & {\bf 4.1} & {\bf 8.3} & 4.3 & 8.6 \\
\bottomrule
\end{tabular}
\end{table}

\begin{table}[tb]
\vspace{-6pt}
\caption{\small Character error rates [\%] of text transcription.}
\label{table:cer_text}
\centering
\footnotesize
\begin{tabular}{lcccc}
\toprule
& eval1 & eval2 & eval3 & JSUT \\
\midrule
\textbf{Whisper} & 18.9 & 17.2 & 15.1 & 8.3 \\
\midrule
\textbf{TT-only} & 4.6 & 5.4 & 5.6 & 17.8 \\
\textbf{MT} & 6.3 & 6.8 & 7.8 & 22.6 \\
\bottomrule
\end{tabular}
\vspace*{-5mm}
\end{table}

\begin{table}[tb]
\vspace{-6pt}%
\caption{\small Mora-label error rates [\%] of each task combination. ``noncore'' shows whether the noncore subset was used in the training.}
\label{table:task_comparison}
\centering
\footnotesize
\begin{tabular}{lccccc}
\toprule
& noncore & eval1 & eval2 & eval3 & JSUT \\
\midrule
\textbf{PA-only} & -- & 11.5 & 11.9 & 13.5 & 13.2 \\
\textbf{PA + TT} & \ding{52} & 7.7  & 8.1 & 9.5 & 10.1 \\
\textbf{PA + \boldFzero}  & -- & 11.8 & 11.9 & 14.0 & 13.7 \\
\textbf{PA + \boldFzero}  & \ding{52} & 11.8 & 12.4 & 13.9 & 12.5 \\
\textbf{PA + TT + \boldFzero (MT)} & \ding{52} &  7.3 & 7.8 & 9.1 & 9.9 \\
\bottomrule
\end{tabular}
\end{table}

Tables~\ref{table:mler} and \ref{table:cer_text} show the mora-label error rates (MLERs) and CERs of the systems, respectively. MLERs were computed both with and without accent errors.
We observed that the generic multilingual speech recognizer (Whisper) was not suitable for our phonemic transcription task.
Even though our final metrics for Whisper were optimistic, computed by choosing the optimal pronunciations based on the reference labels, the results on the CSJ evaluation sets were not competitive with our tailored model.
As discussed previously, this is due to the normalization effect from the language model.
Table~\ref{table:cer_text} shows that Whisper was also not competitive in CER.
This was because CSJ utterances contain a lot of speaker errors, which Whisper tended to ignore.
On the other hand, for JSUT, since this dataset contains read speech without speaker errors, Whisper worked almost perfectly.
In that case, if we could choose the pronunciation of each word optimally, the MLER can be very low (6.8\%).

The Multipa results showed the difficulty of mapping phonetic transcriptions to language-specific phonemic transcriptions.
Spontaneous speech exhibits wide variation in its phonetic realization, making it difficult to recover the phonemic labels as accurately as native speakers do.
This shows that the generic multilingual phonetic transcriber is also not relevant for obtaining phonemic transcriptions.

Comparing PA-only and MT, it was shown that training with the noncore subset was important.
Our method could successfully leverage the noncore subset which lacks accent annotations.
The improvements made by MT were shown to be transferrable to the out-of-domain JSUT task where MLERs reduced from 13.2\% to 9.9\%.

From the results of the decoding methods (``MT+Cond'' and ``MT+LF''), it was shown that the lattice fusion was effective.
The outputs of ``MT+Cond'' suggested that explicit conditioning is fragile if the recognizer cannot recognize textual representation of errors.
We observed that neither decoding method was effective in the JSUT task, due to the inaccuracy of TT outputs used to compute TT lattices (see Table~\ref{table:cer_text}).
This issue was mitigated by introducing an external speech recognizer as a source of TT lattices.
As shown in Table \ref{table:mler}, ``MT+LF with Whisper'' achieved the best result for the JSUT task. Similarly, since our ``TT-only'' model was better than Whisper for CSJ tasks, ``MT+LF with TT-only'' showed the best results for the CSJ tasks.
Thus, we confirmed that ``MT+LF'' approach could further be enhanced if we had an accurate TT recognizer.

To further analyze how each task contributes in multitask training, comparative experiments over each combination of auxiliary task were conducted.
As shown in Table~\ref{table:task_comparison}, most of the advantage of multitask training is attributable to the TT task.
The \Fzero task was not effective if it was only an auxiliary task (PA+\Fzero rows).
However, \Fzero yields a considerable gain when it was combined with TT,
suggesting that \Fzero can be used to regularize the TT task.
Since TT labels can only have limited information on accents, adding \Fzero in conjunction with TT was important to obtain a reliable recognizer that can also leverage prosodic information.

%% file: conclusions.tex
\section{Conclusions}

In this paper, we proposed and evaluated methods for building tailored speech recognizers for Japanese speaking assessment.
The proposed recognizer is equipped with phonetic alphabet and text token decoders, and the phonetic alphabet results are decoded using a lattice fusion technique that integrates the recognition results from text token decoders with using a pronunciation dictionary.
The recognizer is trained on CSJ, which contains phonetic annotations of mispronunciations, hesitations and accents.
Our model training uses a multitask learning scheme consisting of phonetic transcription prediction, text-token prediction, and \Fzero pattern classification.

Our results indicate that the use of multitask learning and lattice fusion was essential to build an accurate phonemic recognizer.
We also verified that a general purpose speech recognizer is not very suitable for building a speaking assessment recognizer since it tends to correct speaker errors.
The extra subtasks, text-token prediction and \Fzero classification, were shown to both contribute to improving estimation accuracy of the main phonemic transcription task.

In this paper, we only focused on the performance of decoders after a single phase of training.
However, given the high accuracy of our models, an iterative training scheme that completes the unannotated part of the training dataset by using self-generated labels is also promising.
Developing a training method with generated labels is a promising future research direction.